\documentclass{article}

\usepackage{PRIMEarxiv}

\usepackage[utf8]{inputenc} 
\usepackage[T1]{fontenc}    
\usepackage{hyperref}       
\usepackage{url}            
\usepackage{booktabs}       
\usepackage{amsfonts}       
\usepackage{nicefrac}       
\usepackage{microtype}      
\usepackage{lipsum}
\usepackage{fancyhdr}       
\usepackage{graphicx}       
\graphicspath{{media/}}     
\usepackage{cite}
\usepackage{amsmath,amssymb,amsfonts}
\usepackage{algorithmic}
\usepackage{textcomp}
\usepackage{xcolor}
\usepackage{amsmath,bm}
\usepackage{microtype}
\usepackage{graphicx}
\usepackage{booktabs} %
\usepackage[caption=false]{subfig}

\title{TransAdapter: Vision Transformer for Feature-Centric Unsupervised Domain Adaptation
}

\author{
  \And
  A. Enes Doruk \\
  Department of Artificial Intelligence and Data Eng. \\
  Ozyegin University \\
  Istanbul, Turkiye\\
  \texttt{enes.doruk@ozu.edu.tr} \\
   \And
  Erhan Oztop \\
  Department of Artificial Intelligence and Data Eng. \\
  Ozyegin University \\
  Istanbul, Turkiye\\
  \texttt{erhan.oztop@ozyegin.edu.tr} \\
   \And
  Hasan F. Ates \\
  Department of Artificial Intelligence and Data Eng. \\
  Ozyegin University \\
  Istanbul, Turkiye\\
  \texttt{hasan.ates@ozyegin.edu.tr} \\
}

\begin{document}
\maketitle

\begin{abstract}
Unsupervised Domain Adaptation (UDA) aims to utilize labeled data from a source domain to solve tasks in an unlabeled target domain, often hindered by significant domain gaps. Traditional CNN-based methods struggle to fully capture complex domain relationships, motivating the shift to vision transformers like the Swin Transformer, which excel in modeling both local and global dependencies. In this work, we propose a novel UDA approach leveraging the Swin Transformer with three key modules. A Graph Domain Discriminator enhances domain alignment by capturing inter-pixel correlations through graph convolutions and entropy-based attention differentiation. An Adaptive Double Attention module combines Windows and Shifted Windows attention with dynamic reweighting to align long-range and local features effectively. Finally, a Cross-Feature Transform modifies Swin Transformer blocks to improve generalization across domains. Extensive benchmarks confirm the state-of-the-art performance of our versatile method, which requires no task-specific alignment modules, establishing its adaptability to diverse applications. our code available at \href{https://github.com/enesdoruk/TransAdapter}{enesdoruk/TransAdapter}.

\end{abstract}

\keywords{Domain Adaptation \and Unsupervised learning \and Transformer}

\section{Introduction}

Deep neural networks (DNNs) have significantly advanced computer vision, excelling in diverse tasks \cite{wang2022, qian2021, yiqi2022, tan2019, chen2021b, jiang2021}. However, their dependence on large labeled datasets imposes high costs and time constraints \cite{csurka2017, zhao2020, zhang2020, oza2021}. Unsupervised Domain Adaptation (UDA) addresses this challenge by enabling knowledge transfer from labeled source domains to unlabeled target domains, tackling domain shifts \cite{bousmalis2017, kuroki2019, wilson2020, vs2021}.

Traditional UDA methods, relying on Convolutional Neural Networks (CNNs), learn domain-invariant features to reduce domain discrepancies through adversarial training and feature normalization \cite{kang2019, zhang2019, jiang2020, li2021b}. However, CNNs struggle with complex domain shifts and long-range dependencies, limiting cross-domain generalization \cite{morerio2020, jiang2020}.

Transformers, widely adopted in NLP \cite{vaswani2017, devlin2018} and computer vision \cite{dosovitskiy2020, han2020, he2021, khan2021}, offer revolutionary feature learning capabilities. The Swin Transformer \cite{liu2021}, known for its hierarchical structure and shift-window mechanism, achieves success but struggles with long-range dependencies due to its localized attention. This limitation, critical for UDA, along with its reliance on large-scale pretraining and fixed partitioning, hampers generalization to domain-specific nuances and significant shifts.

To overcome these limitations, this work introduces TransAdapter, a novel framework for UDA that enhances the Swin Transformer by integrating three innovative modules: Graph Domain Discriminator, Adaptive Double Attention module, and Cross Feature Transform module. These modules address traditional limitations by improving feature alignment and enhancing generalization across domains.

Contributions of this paper are summarized as follows:
\begin{itemize} 
\item Our Graph Domain Discriminator: Unlike CNNs, which focus on local spatial correlations, this discriminator uses a Graph Convolutional Network (GCN) to model non-Euclidean relationships between features. By employing an adjacency matrix based on cosine similarity, it captures both shallow and deep feature dependencies in a scale-invariant manner. This enables holistic domain alignment at individual feature and relational levels, improving feature transferability across domains.

\item The Adaptive Double Attention module simultaneously processes window and shifted window attention features, effectively capturing long-range dependencies crucial for robust domain adaptation. An attention reweighting mechanism emphasizes significant features, while an entropy matrix generated using Graph Domain Discriminator features guides domain alignment. This matrix highlights domain-invariant patterns, enhancing feature transferability.

\item The Cross-Feature Transform module applies dynamic, bidirectional feature transformations between source and target domains using gating attention. By balancing directional contributions through cross-attention and gating mechanisms, it bridges domain gaps effectively. Pairwise feature distances combined with gating outputs ensure adaptive feature alignment and improved generalization across datasets.

\item We utilize CutMix and MixUp as pixelwise feature transform strategies on source data, guided by high-confidence pseudo-labels generated from a Swin-Base model. 
\end{itemize}

Integrating these modules within the Swin Transformer, TransAdapter effectively addresses UDA challenges by leveraging transformers to handle domain shifts, long-range dependencies, and domain-specific nuances, setting a new standard for domain adaptation in vision tasks.

\section{Related Work}
\subsection{Unsupervised Domain Adaptation (UDA) and Transfer Learning}
Unsupervised Domain Adaptation (UDA) within transfer learning focuses on learning transferable knowledge that generalizes across domains with varying data distributions. The primary challenge is addressing domain shift—the discrepancy in probability distributions between source and target domains. Early UDA methods, such as Deep Domain Confusion (DDC), minimized the maximum mean discrepancy (MMD) to learn domain-invariant characteristics~\cite{ddc}. Long et al.~\cite{long_mmd} enhanced this by embedding hidden representations in a reproducing kernel Hilbert space (RKHS) and applying a multiple-kernel variant of MMD for more effective domain distance measurement. Hidden representations refer to the activations within layers of a neural network, capturing hierarchical features of input data.Long et al.\cite{long_joint_mmd} later aligned the joint distributions of multiple domain-specific layers across domains using a joint maximum mean discrepancy (JMMD) metric. Adversarial learning methods, inspired by GANs, also gained popularity.

\subsection{UDA with Vision Transformers}
Herath et al.~\cite{herath2023} proposed an energy-based self-training and normalization approach for UDA, leveraging energy-based learning to improve instance selection on unlabeled target domains. Their method aligns and normalizes energy scores to learn domain-invariant representations, achieving superior performance on benchmarks like DomainNet, Office-Home, and VISDA2017. Sanyal et al.~\cite{sanyal2023} introduced Domain-Specificity Inducing Transformers for Source-Free Domain Adaptation, using vision transformers in privacy-oriented source-free settings. Their approach, leveraging Domain-Representative Inputs and novel domain tokens, achieves state-of-the-art performance across single-source, multi-source, and multi-target benchmarks. Alijani et al.\cite{alijani2024} categorized vision transformers' role in domain adaptation and generalization into feature-level, instance-level, model-level, and hybrid adaptations, highlighting their robustness to distribution shifts. Du et al.\cite{du2024} introduced Domain-Agnostic Mutual Prompting for UDA, leveraging pre-trained vision-language models to address limitations of traditional UDA methods by utilizing rich semantic knowledge and handling complex domain shifts effectively. These studies highlight the growing importance of vision transformers in UDA, offering innovative solutions for domain shifts and enhancing generalization across domains.

\begin{figure*}[t]
\centering
\includegraphics[width=1 \linewidth]{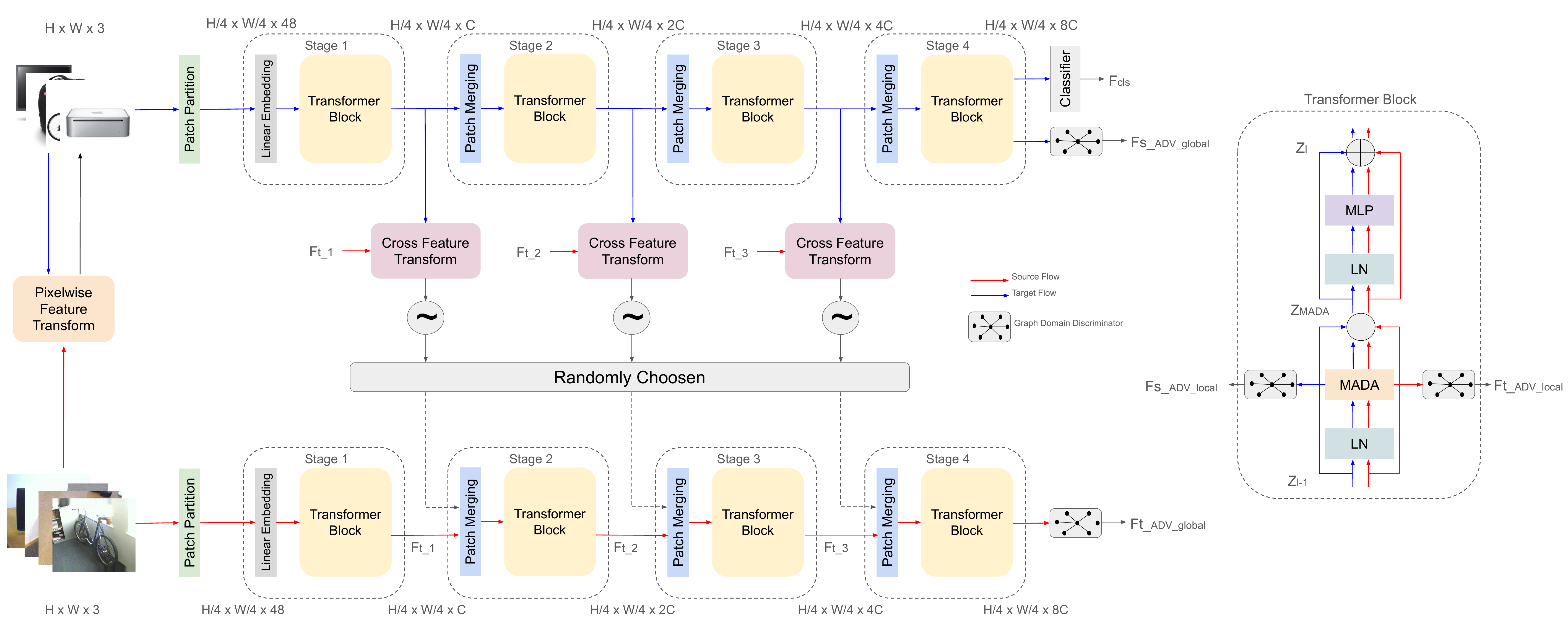}
\caption{ The architecture of the proposed TransAdapter; MADA is multi-head adaptive double attention module, respectively.}
\label{fig:TTP}
\end{figure*}

\section{Method}
Before demonstrating how our method reduces the domain gap in domain-adaptive vision transformer, we first outline the problem formulation. Let \( D_s = \{ (X_s, C_s) \} \) represent a set of \( N_s \) labeled samples in the source domain, where \( X_s = \{ x_i^s \}_{i=1}^{N_s} \) are the input samples and \( C_s = \{ c_i^s \}_{i=1}^{N_s} \) are their corresponding class labels. In the target domain, we have \( D_t = \{ X_t \} \), consisting of \( N_t \) unlabeled samples, \( X_t = \{ x_j^t \}_{j=1}^{N_t} \), with no labels. The objective in this unsupervised domain adaptation task is to develop a classifier that generalizes across domains using both \( D_s \) and \( D_t \).

\subsection{Adaptive Double Attention}
\begin{figure*}[t]
\centering
\includegraphics[width= 0.9 \textwidth]{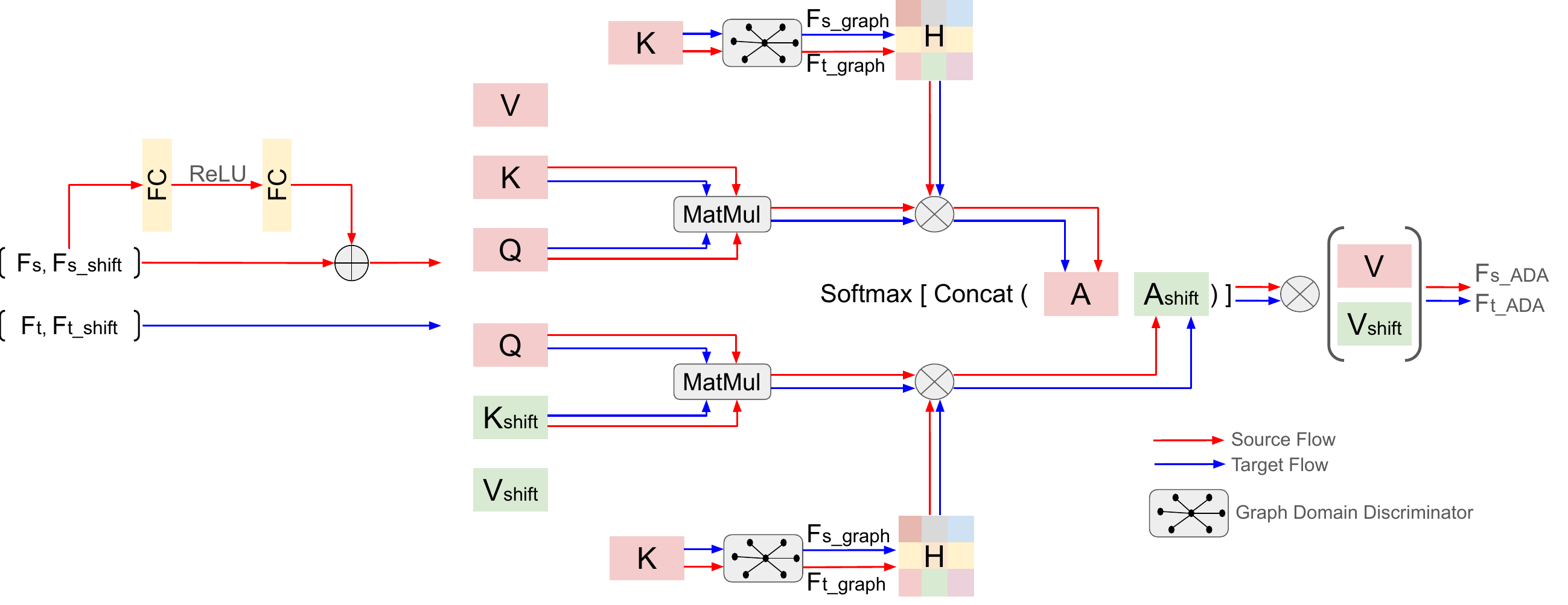}
\caption{The architecture of the Adaptive Double Attention (ADA) module is depicted. Here, \( KQV \) represents the standard window attention features, while \( K_{\text{shift}}Q_{\text{shift}}V_{\text{shift}} \) corresponds to the shifted window attention features. Additionally, \( H \) denotes the entropy-based attention matrix, which is constructed using features derived from the graph domain discriminator.}
\label{fig:ADA}
\end{figure*}

The Adaptive Double Attention (ADA) module, shown in Figure \ref{fig:ADA}, introduces an entropy-guided mechanism to address domain alignment challenges in unsupervised domain adaptation (UDA), particularly for long-range dependencies. While existing architectures, such as vanilla ViTs and Swin Transformers, model local and global dependencies, they lack effective mechanisms for domain alignment. ADA resolves this by integrating feature correction, double attention mechanisms, and entropy-guided reweighting, dynamically aligning source and target domain representations.

A key feature of ADA is entropy-guided reweighting, integrated directly into the attention process. The entropy, calculated from outputs of a graph domain discriminator, prioritizes transferable features while suppressing domain-specific ones:

\begin{equation}
H(F_{\text{graph}}) = - \sum_{i} F_{\text{graph}} \log(F_{\text{graph}}).
\end{equation}

The graph domain discriminator processes key and shifted key features, generating outputs \( F_{s_{\text{graph}}} \) and \( F_{t_{\text{graph}}} \). Lower entropy indicates better alignment, while higher entropy signals domain-specific noise. Entropy values dynamically reweight the attention scores:

\begin{equation}
\begin{aligned}
A &= \frac{Q K^T}{\sqrt{d}} \odot H(F_{\text{graph}}), \\
A_{\text{shift}} &= \frac{Q K_{\text{shift}}^T}{\sqrt{d}} \odot H(F_{\text{graph}}).
\end{aligned}
\end{equation}

The reweighted scores are concatenated, normalized with the softmax function, and combined with the value vectors:

\begin{equation}
MADA = \text{Softmax}(\text{Concat}(A, A_{\text{shift}})) \times [V; V_{\text{shift}}].
\end{equation}

To further minimize domain discrepancies, ADA employs a feature correction step before attention. Inspired by prior work, this correction block modifies target features by incorporating a correction term, implemented using two fully connected layers with ReLU activations. This ensures harmonized inputs for attention mechanisms. Window attention captures fine-grained spatial details within local regions, while shifted window attention models global dependencies. ADA integrates these mechanisms through cross-attention, where queries from window attention interact with keys from shifted attention, unifying local and global dependencies.

The final output of the adaptive attention mechanism, \( Z_{\text{MADA}} \), is computed as follows:

\begin{equation}
\begin{aligned}
Z_{\text{MADA}} &= MADA(\text{LN}(Z_{l-1})) + Z_{l-1}, \\
Z_l &= \text{MLP}(\text{LN}(Z_{\text{MADA}})) + Z_{\text{MADA}}.
\end{aligned}
\end{equation}

Here, \( Z_{l-1} \) is the input to the transformer block, \( Z_{\text{MADA}} \) is the adaptive attention output, and \( Z_l \) is the final block output. Residual connections and layer normalization stabilize learning and ensure efficient gradient flow.

By combining entropy-guided reweighting with dual attention mechanisms, the ADA module prioritizes transferable features, aligning long-range dependencies effectively. This robust approach addresses the limitations of existing architectures like Swin Transformers, enhancing domain alignment and improving generalization across diverse domains.

\subsection{Graph Domain Discriminator}
Graph convolutions in the domain discriminator explicitly model inter-sample relationships, critical for domain alignment. Unlike methods like DANN \cite{ganin2015unsupervised}, which process samples independently, graph convolutions operate on an adjacency matrix encoding pairwise relationships between source and target samples. This allows the \textbf{Graph Domain Discriminator (GDD)} to leverage global and local topological dependencies, enabling a nuanced understanding of domain shifts.

\begin{figure}[ht]
\centering
\includegraphics[width= 0.6 \textwidth]{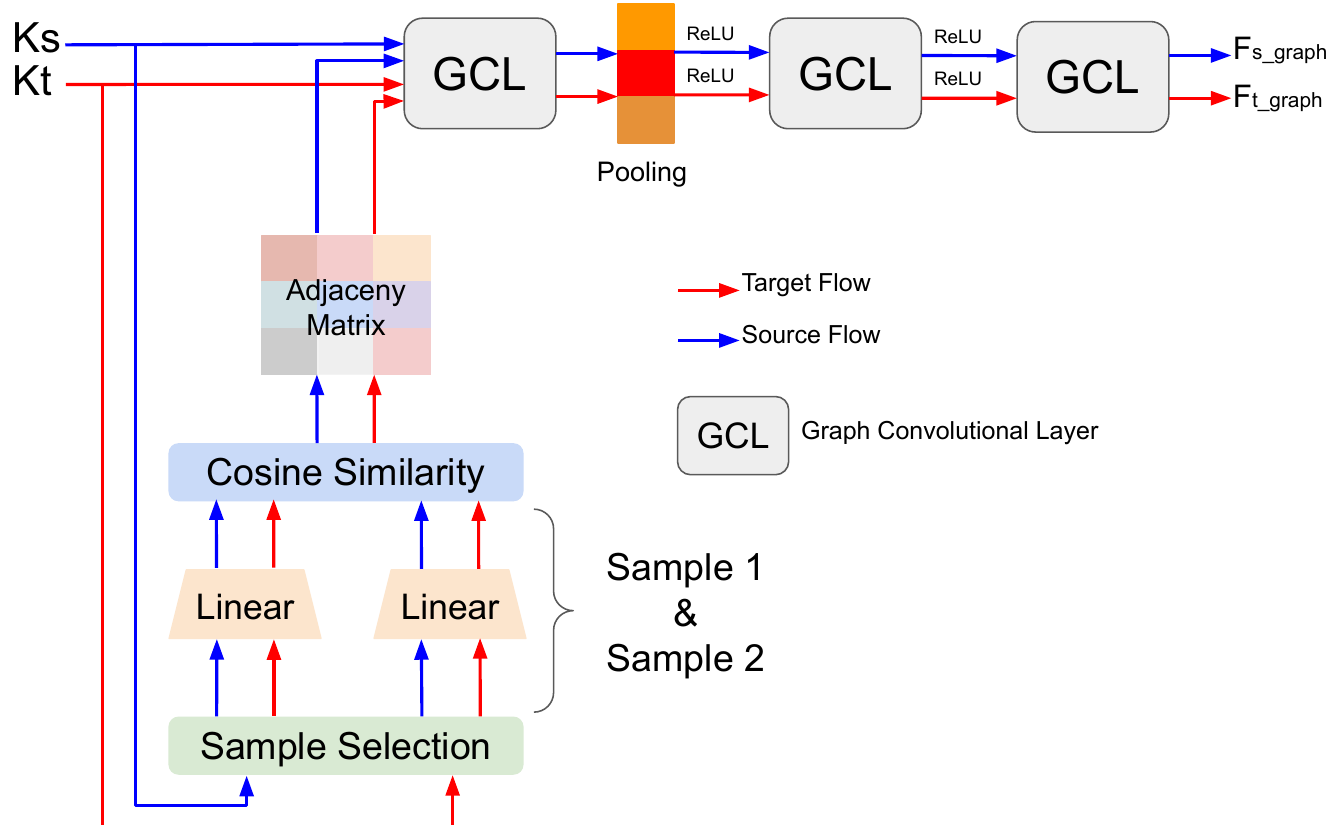}
\caption{The architecture of the Graph Domain Discriminator uses \(K_s\) and \(K_t\) to represent source and target key features of MADA, respectively.}
\label{fig:GDD}
\end{figure}

The adjacency matrix, central to GDD, is constructed using cosine similarities between learnable projections of sample features:

\begin{equation}
    \frac{\bm{P}(\mathbf{x}_i) \cdot \bm{P}(\mathbf{x}_j)}{|\bm{P}(\mathbf{x}_i)| |\bm{P}(\mathbf{x}_j)|},
\end{equation}

where \(\bm{P}(\mathbf{x}_i)\) and \(\bm{P}(\mathbf{x}_j)\) are the projected features of samples \(i\) and \(j\). This matrix enables GDD to capture inter-sample dependencies across domains, which are essential for modeling domain shifts that often involve subtle feature variations. By propagating relational information across samples, the adjacency matrix allows GDD to consider both individual domain-specific characteristics and structural interactions, fostering a more comprehensive domain alignment.

The GDD employs three graph convolutional layers with ReLU activation to aggregate information from sample neighbors, enriching domain-shared feature representations. A pooling operation after the first layer reduces dimensionality while emphasizing salient features. By leveraging local features from the \((N-2)\)-th transformer block and global features from the \(N\)-th block, GDD achieves hierarchical alignment of domain representations.

To promote domain invariance, a Gradient Reversal Layer (GRL) is incorporated after the graph convolutional layers, facilitating a min-max optimization process. This setup enables the domain discriminator to minimize domain-specific biases while guiding the feature extractor to generate domain-invariant features. By simultaneously modeling global feature distributions and fine-grained inter-domain relationships, GDD achieves robust domain alignment, improving the adaptability of the shared feature space.

\subsection{Cross Feature Transform}
\begin{figure}[t]
\centering
\includegraphics[width= 0.35 \textwidth]{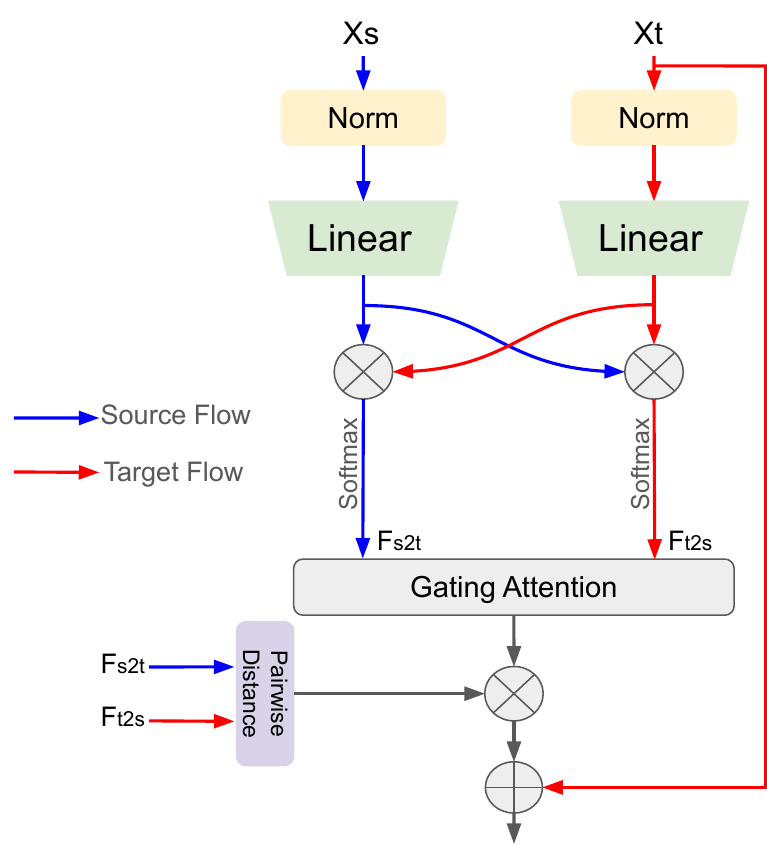}
\caption{The architecture of Cross Feature Transform (CFT) module. $X_s$ and $X_t$ represents source and target feature, respectively.
}
\vspace{-3mm}
\label{fig:CFT}
\end{figure}

The proposed Cross Feature Transform (CFT) module enhances domain adaptation within the Transformer architecture by facilitating effective feature alignment between source and target domains. Unlike static methods, the CFT module is applied dynamically after a randomly selected transformer block in each iteration, providing a robust feature transformation approach and reducing the likelihood of overfitting \cite{sun2022safe}. The general architecture of the CFT module is illustrated in Figure~\ref{fig:CFT}.

Central to the CFT module are bidirectional cross-attention mechanisms, which optimize feature transferability between domains, enabling implicit mixing of features. This enhances the model's ability to learn domain-invariant representations, thereby improving generalization to the target domain \cite{wang2022domain}. The computation of source-to-target attention features \(F_{s2t}\) and target-to-source attention features \(F_{t2s}\) is performed as follows:

\begin{equation}
\begin{aligned}
F_{s2t} = \text{Softmax}\left(f(X_s)^\top g(X_t)\right)\\
F_{t2s} = \text{Softmax}\left(g(X_t)^\top f(X_s)\right)   
\end{aligned}
\end{equation}

To refine feature alignment, the CFT module incorporates a gating mechanism using a learnable parameter \(\mathbf{\gamma}\), balancing contributions from both directions:

\begin{equation}
\text{Attn}_{gating} = (1 - \sigma(\gamma)) \cdot F_{s2t} + \sigma(\gamma) \cdot F_{t2s}
\end{equation}

where \(\sigma(\gamma)\) is the sigmoid function. This adaptive formulation allows prioritization of source-to-target or target-to-source transformations based on data context.

The pairwise distance between features is computed and combined with the gating attention output:

\begin{equation}
F_{out} = \left(\text{Attn}_{gating} \times \|F_{s2t} - F_{t2s}\|_2^2\right) + X_t
\end{equation}

Here, \(\|F_{s2t} - F_{t2s}\|\) represents the pairwise distance, \(\text{Attn}_{gating}\) the gating attention output, and \(X_t\) is the target feature added as a shortcut.

\subsection{Pixel-Wise Feature Transform with Pseudo Labeling}
We employ CutMix \cite{yun2019cutmix} and MixUp \cite{zhang2017mixup} as pixel-wise transformation strategies on raw images to improve feature transferability between domains. Although these methods generally necessitate labeled data, our unsupervised domain adaptation task operates without ground truth labels in the target domain. To tackle this issue, we generate pseudo-labels for the target data using a Swin-Base model trained on the source domain. To reduce noise in these pseudo-labels, we implement a confidence threshold based on the model’s accuracy, retaining only predictions that exceed this threshold for the transformation operations. These transformation are applied solely to the source data, as our network incorporates a Cross Feature Transform (CFT) module that enhances feature transferability between domains, thus diminishing the necessity for direct transformation on the target data. The pixel-wise CutMix and MixUp operations, guided by high-confidence pseudo-labels, are shown in Figure~\ref{fig:TTP}.

\begin{figure*}[ht]
\centering
\scalebox{0.8}{
\subfloat[$Swin-B$]{%
  \includegraphics[width=0.25\textwidth,keepaspectratio]{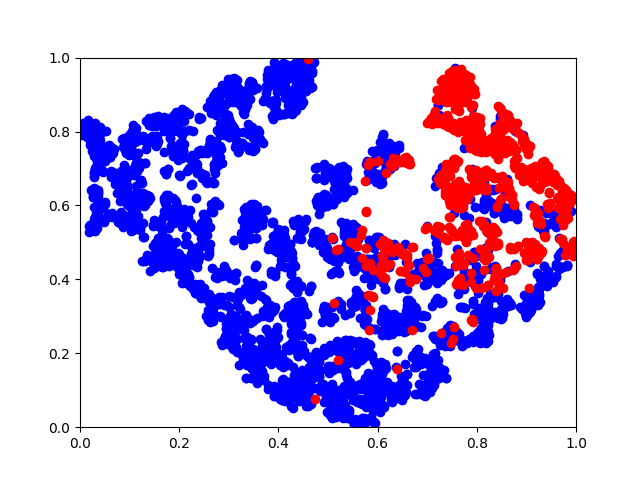}
}
\subfloat[$+GDD$]{%
  \includegraphics[width=0.25\textwidth,keepaspectratio]{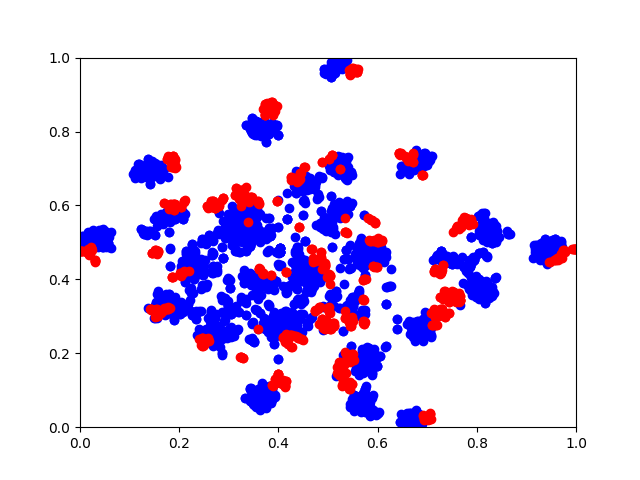}
}
\subfloat[$+Pixelwise Transform$]{%
  \includegraphics[width=0.25\textwidth,keepaspectratio]{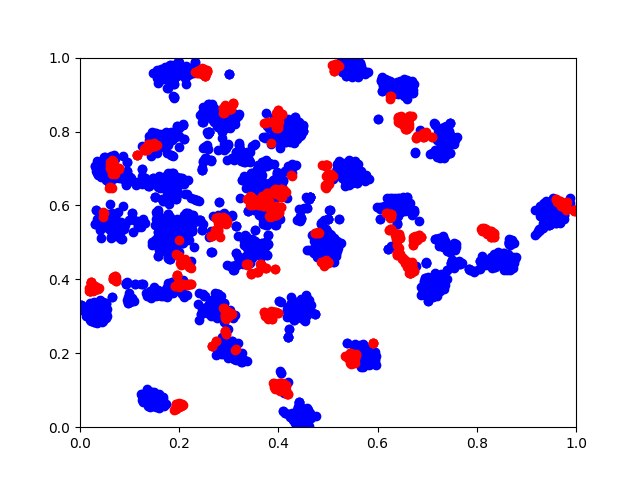}
}
\subfloat[$+CFT$]{%
  \includegraphics[width=0.25\textwidth,keepaspectratio]{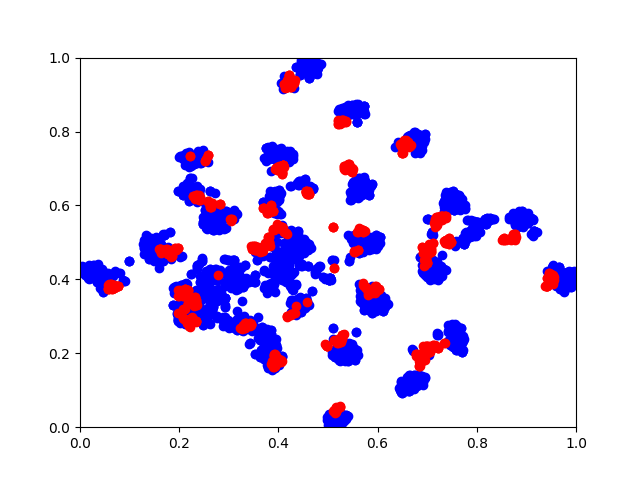}
}

\subfloat[$+ADA (TransAdapter)$]{%
  \includegraphics[width=0.25\textwidth,keepaspectratio]{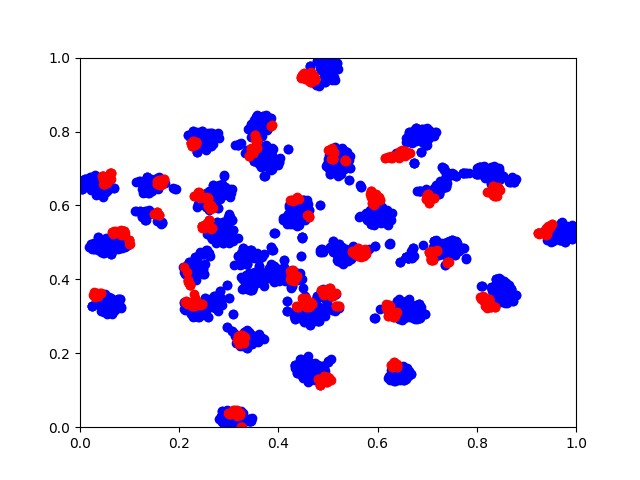}
}
}
\caption{t-SNE visualization of Office-Home dataset, where red and blue points indicate the source and the target domain, 
            ”-B” indicates that the backbone is Base, respectively.}
\label{fig:TSNE}
\end{figure*}

\section{Experiments}
\subsection{Datasets}
The \textbf{Office-31} dataset \cite{saenko2010adapting} contains 4,652 images across 31 categories from three domains: Amazon (A), DSLR (D), and Webcam (W). Images were sourced from Amazon.com or captured in office settings using a DSLR or webcam.

The \textbf{Office-Home} dataset \cite{venkateswara2017deep} includes four domains: Artistic (Ar), Clip Art (Cl), Product (Pr), and Real-World (Rw), with 65 categories per domain, offering diverse evaluation scenarios.

The \textbf{VisDA-2017} dataset \cite{peng2017visda}, designed for synthesis-to-real tasks, includes 12 categories. The source domain contains 152,397 synthetic renderings, while the target domain has 55,388 real-world images.

The \textbf{DomainNet} dataset \cite{peng2019moment}, the largest UDA benchmark, comprises approximately 0.6 million images from six domains: Clipart (Clp), Infograph (Inf), Painting (Pnt), Quickdraw (Qdr), Real (Rel), and Sketch (Skt), covering 345 categories for challenging multi-source and single-source adaptation tasks.

\begin{table}[ht]
    \caption{Accuracy $(\%)$ on the Office-Home dataset. ”-S” and
            ”-B” indicates that the backbone is Small and Base, respectively. The best performance is marked as \textbf{bold}.}
    \label{tab:ofhome}
    \centering
    \resizebox{17cm}{!}{%
    \begin{tabular}{ l | @{}l @{}l @{}l @{}l @{}l @{}l @{}l @{}l @{}l @{}l @{}l @{}l |  l }
    \hline
         \textbf{Method}  & $A \rightarrow CA$ & $ \rightarrow PA$ & $ \rightarrow RC$ & $ \rightarrow AC$ & $ \rightarrow PC$ & $ \rightarrow RP$ & $ \rightarrow AP$ &
         $ \rightarrow CP$ & $ \rightarrow RR$ & $ \rightarrow AR$ & 
         $ \rightarrow CR$ & $ \rightarrow P$ & Avg \\ [0.5ex] 

         \hline \hline
         \multicolumn{14}{c}{\textbf{ResNet} Backbone} \\ \hline
         ResNet \cite{he2016deep}  & $34.9$ & $50.0$ & $58.0$ & $37.4$ & $41.9$ & $46.2$ & $38.5$ & $31.2$ & $60.4$ & $53.9$ & $41.2$ & $59.9$ & $46.1$ \\ 
         DAN \cite{long2015learning}  & $43.6$ & $57.0$ & $67.9$ & $45.8$ & $56.5$ & $60.4$ & $44.0$ & $43.6$ & $67.7$ & $63.1$ & $51.5$ & $74.3$ & $56.3$ \\ 
         RevGrad \cite{ganin2015unsupervised}  & $45.6$ & $59.3$ & $70.1$ & $47.0$ & $58.5$ & $60.9$ & $46.1$ & $43.7$ & $68.5$ & $63.2$ & $51.8$ & $76.8$ & $57.6$ \\ 
         SHOT \cite{liang2020we}  & $57.1$ & $78.1$ & $81.5$ & $68.0$ & $78.2$ & $78.1$ & $67.4$ & $54.9$ & $82.2$ & $73.3$ & $58.8$ & $84.3$ & $71.8$ \\ \hline
         
         \multicolumn{14}{c}{\textbf{DeiT} Backbone} \\ \hline 
         DeiT-B \cite{touvron2021training} & $61.8$ & $79.5$ & $84.3$ & $75.4$ & $78.8$ & $81.2$ & $72.8$ & $55.7$ & $84.4$ & $78.3$ & $59.3$ & $86.0$ & $74.8$ \\ 
         CDTrans-B \cite{xu2021cdtrans} & $68.8$ & $85.0$ & $86.9$ & $81.5$ & $87.1$ & $87.3$ & $79.6$ & $63.3$ & $88.2$ & $82.0$ & $66.0$ & $90.6$ & $80.5$ \\ 
         DeiT-S \cite{touvron2021training} & $54.4$ & $73.8$ & $79.9$ & $68.6$ & $72.6$ & $75.1$ & $63.6$ & $50.2$ & $80.0$ & $73.6$ & $55.2$ & $82.2$ & $69.1$ \\
         WinTR-S \cite{ma2021exploiting} & $65.3$ & $84.1$ & $85.0$ & $76.8$ & $84.5$ & $84.4$ & $73.4$  & $60.0$ & $85.7$ & $77.2$ & $63.1$ & $86.8$ & $77.2$ \\  \hline
         
         \multicolumn{14}{c}{\textbf{Swin} Backbone} \\ \hline
         Swin-S \cite{liu2021swin} & $59.2$ & $75.3$ & $78.5$ & $71.8$ & $74.6$ & $76.7$ & $69.4$ & $54.1$ & $79.5$ & $74.1$ & $59.6$ & $79.8$ & $76.1$ \\
         \textbf{TransAdapter-S (ours)} & $74.2$ & $89.6$ & $90.2$ & $88.9$ & $\textbf{91.3}$ & $88.4$ & $87.1$ & $77.5$ & $90.8$ & $85.2$ & $78.6$ & $90.7$ & $86.3$ \\
         Swin-B \cite{liu2021swin} & $64.5$ & $84.8$ & $87.6$ & $82.2$ & $84.6$ & $86.7$ & $78.8$ & $60.3$ & $88.9$ & $82.8$ & $65.3$ & $89.6$ & $79.7$ \\ 
         BCAT \cite{wang2022domain} & $75.3$ & $90.0$ & $92.9$ & $88.6$ & $90.3$ & $\textbf{92.7}$ & $87.4$ & $73.7$ & $92.5$ & $86.7$ & $75.4$ & $93.5$ & $86.6$\\
         PMTrans-B \cite{zhu2023patch} & $\textbf{81.3}$ & $92.9$ & $92.8$ & $88.4$ & $93.4$ & $93.2$ & $87.9$ & $\textbf{80.4}$ & $\textbf{93.0}$ & $89.0$ & $80.9$ & $\textbf{94.8}$ & $89.0$\\ 
         \textbf{TransAdapter-B (ours)} & $78.6$ & $\textbf{92.8}$ & $\textbf{93.1}$ & $\textbf{93.5}$ & $\textbf{94.3}$ & $92.6$ & $\textbf{91.3}$ & $79.6$ & $92.8$ & $\textbf{89.3}$ & $\textbf{81.1}$ & $92.8$ & \colorbox{lightgray}{\textbf{89.4}} \\ \hline
         \multicolumn{14}{c}{\textbf{ViT} Backbone} \\ \hline
         ViT-B \cite{alexey2020image} & $66.2$ & $84.3$ & $86.6$ & $77.9$ & $83.3$ & $84.3$ & $76.0$ & $62.7$ & $88.7$ & $80.1$ & $66.2$ & $88.7$ & $78.7$ \\ 
         TVT-B \cite{yang2023tvt} & $74.9$ & $86.8$ & $89.5$ & $82.8$ & $88.0$ & $88.3$ & $79.8$ & $71.9$ & $90.1$ & $85.5$ & $74.6$ & $90.6$ & $83.6$ \\
         SSRT-B \cite{sun2022safe} & $75.1$ & $88.9$ & $91.0$ & $85.1$ & $88.2$ & $89.9$ & $85.0$ & $74.2$ & $91.2$ & $85.7$ & $78.5$ & $91.7$ & $85.4$ \\
         PDA-B \cite{bai2024prompt} & $73.5$ & $91.4$ & $91.3$ & $86.0$ & $91.6$ & $91.5$ & $86.0$ & $73.5$ & $91.7$ & $86.4$ & $73.0$ & $92.4$ & $85.7$ \\
         PADCLIP-B \cite{lai2023padclip} & $76.4$ & $90.6$ & $90.8$ & $86.7$ & $92.3$ & $92.0$ & $86.0$ & $74.5$ & $91.5$ & $86.9$ & $79.1$ & $92.1$ & $86.7$ \\
         \hline
         
    \end{tabular} 
    }
\end{table}

\subsection{Implementation Details}
For all domain adaptation (DA) tasks, we utilize the Swin model, pretrained on the ImageNet dataset~\cite{deng2009imagenet}, as the backbone network in our proposed TransAdapter framework. Additionally, we construct two model variants: \textbf{TransAdapter-S} and \textbf{TransAdapter-B}, derived respectively from Swin-S and Swin-B backbones, integrating 12 dual transformer blocks from their corresponding Swin architectures within the TransAdapter framework. The models are optimized using the Stochastic Gradient Descent (SGD) algorithm~\cite{bottou2010large}, with a momentum of 0.9 and a weight decay parameter of $1 \times 10^{-3}$. We employ a base learning rate of $1 \times 10^{-2}$ for the Office-31 and Office-Home datasets, while a lower learning rate of $1 \times 10^{-3}$ is applied for the VisDA-2017 dataset. The learning rate follows a warmup cosine scheduler, gradually increasing during the initial training phase and subsequently decaying throughout the remaining iterations. Across all datasets, the batch size is consistently set to 32, and the models are trained over $15,000$ iterations. The hyperparameters $\lambda_{\text{local}}$ and $\lambda_{\text{global}}$ in the TransAdapter method are set to $0.1$ and $0.01$, respectively, for all DA tasks, as shown in Equation~\ref{eq:adv_loss}.

\subsection{Objective Function}
The domain adaptive model optimizes a combined objective function comprising cross-entropy loss for classification, local adaptation loss (strong alignment), and global adaptation loss (weak alignment). The classification loss for the labeled source domain is:

\begin{equation}
L_{\text{cls}} = \text{CE}(F_{cls}, y_s),
\end{equation}

where $F_{cls}$ denotes last transformer block output, \( y_s \) is the ground truth for the source domain, and \(\text{CE}\) represents cross-entropy loss.

For adaptation, local and global losses are computed as averages over source and target domains:

\begin{equation}
\begin{aligned}
L_{local} = \frac{1}{2} \big( \text{CE}(F_{ADV\_local}^{src}, \hat{y}^{src}) 
          + \text{CE}(F_{ADV\_local}^{tgt}, \hat{y}^{tgt}) \big) \\
L_{global} = \frac{1}{2} \big( FL(F_{ADV\_global}^{src}, \hat{y}^{src}) 
          + FL(F_{ADV\_global}^{tgt}, \hat{y}^{tgt}) \big)
\end{aligned}
\end{equation}

where \( \hat{y}^{\text{src}} = 1 \) and \( \hat{y}^{\text{tgt}} = 0 \). \( F_{\text{ADV}} \) denotes the output of the second transformer block for local alignment and the output of the last transformer block for global alignment. \( FL(\cdot) \) represents the focal loss function designed to address class imbalance.

The total loss is:

\begin{equation}
\mathcal{L}_{\text{total}} = \lambda_{\text{local}} \mathcal{L}_{\text{local}} + \lambda_{\text{global}} \mathcal{L}_{\text{global}} + \mathcal{L}_{\text{cls}},
\label{eq:adv_loss}
\end{equation}

where \(\lambda_{\text{local}}\) and \(\lambda_{\text{global}}\) are weighting coefficients.

\begin{table*}
\centering
\vspace{-5pt}
\captionsetup{font=small}
\caption{Comparison with SoTA methods on DomainNet. The best performance is marked as \textbf{bold}. ”-B” indicates that the backbone is Base, respectively.}
\resizebox{\linewidth}{!}{ 
\begin{tabular}{c|lllllll||c|lllllll||c|llllllll}
\toprule
MCD& clp& inf& pnt& qdr& rel& skt& Avg&
SWD& clp& inf& pnt& qdr& rel& skt& Avg&
BNM& clp& inf& pnt& qdr& rel& skt& Avg  \\
\hline
\hline
clp&-&15.4&25.5&3.3&44.6&31.2&24.0& 
clp&-&14.7&31.9&10.1&45.3&36.5&27.7& 
clp&-&12.1&33.1&6.2&50.8&40.2&28.5  \\

inf&24.1&-&24.0&1.6&35.2&19.7& 20.9&
inf&22.9&-&24.2&2.5&33.2&21.3&20.0&
inf&26.6&-&28.5&2.4&38.5&18.1&22.8\\

pnt&31.1&14.8&-&1.7&48.1&22.8& 23.7& 
pnt&33.6&15.3&-&4.4&46.1&30.7&26.0&
pnt&39.9&12.2&-&3.4&54.5&36.2&29.2\\

qdr&8.5&2.1&4.6&-&7.9&7.1&6.0& 
qdr&15.5&2.2&6.4&-&11.1&10.2&9.1&
qdr&17.8&1.0&3.6&-&9.2&8.3&8.0\\

rel&39.4&17.8&41.2&1.5&-&25.2&25.0&
real&41.2&18.1&44.2&4.6&-&31.6&27.9&
rel&48.6&13.2&49.7&3.6&-&33.9&29.8 \\

skt&37.3&12.6&27.2&4.1&34.5&-&23.1& 
skt&44.2&15.2&37.3&10.3&44.7&-&30.3&
skt&54.9&12.8&42.3&5.4&51.3&-&33.3 \\ 

Avg&28.1&12.5&24.5&2.4&34.1&21.2&
\colorbox{lightgray}{20.5}&Avg&31.5&13.1&28.8&6.4&36.1&26.1&
\colorbox{lightgray}{23.6}&Avg&37.6&10.3&31.4&4.2&40.9&27.3& \colorbox{lightgray}{25.3} \\
\hline
\hline
CGDM & clp& inf& pnt& qdr& rel& skt& Avg&
PMTrans& clp& inf& pnt& qdr& rel& skt& Avg&
SCDA & clp& inf& pnt& qdr& rel& skt& Avg  \\
\hline
\hline
clp&-&16.9&35.3&10.8&53.5&36.9&30.7& 
clp&-&34.2&62.7&32.5&79.3&63.7&54.5&
clp&-&18.6&39.3&5.1&55.0&44.1&32.4\\

inf&27.8&-&28.2&4.4&48.2&22.5&26.2&
inf&67.4&-&61.1&22.2&78.0&57.6&57.3&
inf&29.6&-&34.0&1.4&46.3&25.4&27.3 \\ 

pnt&37.7&14.5&-&4.6&59.4&33.5&30.0&
pnt&69.7&33.5&-&23.9&79.8&61.2&53.6&
pnt&44.1&19.0&-&2.6&56.2&42.0&32.8 \\

qdr&14.9&1.5&6.2&-&10.9&10.2&8.7&
qdr&54.6&17.4&38.9&-&49.5&41.0&40.3&
qdr&30.0&4.9&15.0&-&25.4&19.8&19.0\\

rel&49.4&20.8&47.2&4.8&-&38.2&32.0& 
rel&74.1&35.3&70.0&25.4&-&61.1&53.2&
rel&54.0&22.5&51.9&2.3&-&42.5&34.6\\

skt&50.1&16.5&43.7&11.1&55.6&-&35.4& 
skt&73.8&33.0&62.6&30.9&77.5&-& 55.6&
skt&55.6&18.5&44.7&6.4&53.2&-&35.7\\ 

Avg&36.0&14.0&32.1&7.1&45.5&28.3& 
\colorbox{lightgray}{27.2}&Avg&40.4&16.6&34.7&4.3&43.4&32.3& 
\colorbox{lightgray}{28.6}&Avg&42.6&16.7&37.0&3.6&47.2&34.8& \colorbox{lightgray}{30.3}\\
\hline
\hline
CDTrans& clp& inf& pnt& qdr& rel& skt& Avg&
SSRT& clp& inf& pnt& qdr& rel& skt& Avg&
\textbf{TransAdapter-B}& clp& inf& pnt& qdr& rel& skt& Avg\\
\hline
\hline
clp&-&29.4&57.2&26.0&72.6&58.1&48.7&
clp&-&33.8&60.2&19.4&75.8&59.8&49.8&    
clp&-&35.4&63.9&33.7&80.4&65.0&55.8 \\

inf&57.0&-&54.4&12.8&69.5&48.4&48.4&
inf&55.5&-&54.0&9.0&68.2&44.7&46.3&
inf&68.8&-&62.5&23.6&79.4&59.0&58.7 \\ 

pnt&62.9&27.4&-&15.8&72.1&53.9&46.4&
pnt&61.7&28.5&-&8.4&71.4&55.2&45.0&     
pnt&70.8&34.6&-&25.0&80.9&62.3&54.7 \\

qdr&44.6&8.9&29.0&-&42.6&28.5&30.7&
qdr&42.5&8.8&24.2&-&37.6&33.6&29.3&  
qdr&55.8&18.6&40.1&-&50.7&42.2&41.5 \\

rel&66.2&31.0&61.5&16.2&-&52.9&45.6&
rel&69.9&37.1&66.0&10.1&-&58.9& 48.4&      
rel&75.3&36.5&71.2&26.6&-&62.3&54.4 \\

skt&69.0&29.6&59.0&27.2&72.5&-&51.5&
skt&70.6&32.8&62.2&21.7&73.2&-&52.1&   
skt&75.2&34.4&64.0&32.3&78.9&-& 57.0\\ 

Avg&59.9&25.3&52.2&19.6&65.9&48.4&
\colorbox{lightgray}{45.2}&Avg&60.0&28.2&53.3&13.7&65.3&50.4& 
\colorbox{lightgray}{45.2}&Avg&67.9&30.7&59.1&27.0&72.8&56.9&\colorbox{lightgray}{\textbf{53.7}}  \\
\bottomrule
\end{tabular}}
\vspace{-8pt}
\label{tab:domainnet}
\end{table*}

\subsection{Ablation Study}
Table \ref{tab:abb} presents the ablation study results, demonstrating the impact of each proposed module in our model. Adding the Graph Domain Discriminator (GDA) improves domain alignment by modeling complex feature relationships, resulting in a performance boost across all datasets, with notable gains on VisDA-2017 ($+4.9\%$) and DomainNet ($+6.0\%$). Introducing Pixelwise Transform further enhances performance by leveraging high-confidence pseudo-labels for effective feature augmentation, yielding an additional improvement of $+1.2\%$ to $+2.1\%$ across datasets. The Cross-Feature Transform (CFT) module significantly bridges domain gaps through dynamic, bidirectional feature transformations, leading to remarkable gains, particularly on VisDA-2017 ($+4.8\%$) and DomainNet ($+2.4\%$).

\begin{table}[ht]
\centering
\caption{Ablation study of each module $(\%)$. The best performance is marked as \textbf{bold}. Last row corresponds the proposed model.
            ”-B” indicates that the backbone is Base, respectively.}
\label{tab:abb}
\scalebox{0.7}{
\begin{tabular}{ l |  l l l l }
  \hline
         \textbf{Method}  & Office-31 & Office-Home & VisDA-2017 & DomainNet \\[0.5ex] 
         \hline \hline
         Swin-B \cite{liu2021swin}  &  $89.8$ & $79.7$ & $73.9$ & $41.2$ \\
         CNN Discriminator  &  $90.5$ & $80.7$ & $75.2$ & $43.8$ \\
         +GDD   & $91.7$ & $81.6$ & $78.8$ & $47.2$ \\ 
         +Pixelwise Transform &  $92.9$ & $82.6$ & $79.8$ & $49.3$ \\
         +CFT   & $93.5$ & $84.1$ & $84.6$ & $51.7$ \\ 
         +ADA (TransAdapter)  & $\textbf{95.5}$ & $\textbf{87.5}$ & $\textbf{90.2}$ & $\textbf{53.7}$ \\ \hline 
\end{tabular}}
\vspace{-2mm}
\end{table}

\textbf{CNN vs GCN Based Discriminator.} The results in Table \ref{tab:abb} demonstrate the advantage of GDA, which employs a Graph Convolutional Network (GCN) to model complex, non-Euclidean relationships using an adjacency matrix based on cosine similarity. Unlike CNNs, which focus on local spatial dependencies, GDA captures global context and non-local feature correlations for comprehensive domain alignment. This approach achieves notable improvements on challenging datasets like VisDA-2017 (+3.6\%) and DomainNet (+3.4\%), highlighting GDA's effectiveness in addressing significant domain shifts in diverse and complex scenarios.

\textbf{Visualization.} Figure \ref{fig:TSNE} shows t-SNE visualizations of domain discrepancies using final transformer block features from TransAdapter-B. Adding the GDD on Swin-Base model reduces gaps by modeling complex feature relationships, while Pixelwise Transform enhances alignment through pseudo-label-guided augmentation. The CFT dynamically bridges domain gaps, resulting in more cohesive feature distributions. The complete TransAdapter-B after adding ADA achieves the most compact and well-aligned clusters, effectively minimizing domain discrepancies.

\begin{table}[ht]
    \caption{Accuracy $(\%)$ on the VisDA-2017 dataset. 
            ”-S” and ”-B” indicates that the backbone is Small and Base, respectively. The best performance is marked as \textbf{bold}.}
    \label{tab:visda}
    \centering
    \setlength{\tabcolsep}{2pt} 
    \resizebox{17cm}{!}{%
    \begin{tabular}{ l | l l l l l l l l l l l l | l }
    \hline
         \textbf{Method}  & plane & bcycl & bus & car & house & knife
         & mcycl & person & plant & sktbrd & train & truck & Avg \\ [0.5ex] 
         
         \hline \hline
         
        \multicolumn{14}{c}{\textbf{ResNet} Backbone} \\ \hline
        ResNet \cite{he2016deep}  & $55.1$ & $53.3$ & $61.9$ & $59.1$
                        &  $80.6$ & $17.9$ & $79.7$ & $31.2$
                        & $81.0$ & $26.5$ & $73.5$ 
                        & $8.5$ & $52.4$ \\            
        RevGrad \cite{ganin2015unsupervised}  & $81.9$ & $77.7$ & $82.8$ & $44.3$ 
                                                & $81.2$ & $29.5$ & $65.1$ & $28.6$ 
                                                & $51.9$ & $54.6$ & $82.8$
                                                & $7.8$ & $57.4$   \\ 
        MCD \cite{saito2018maximum}  & $87.0$ & $60.9$ & $83.7$ & $64.0$ 
                                    & $88.9$ & $79.6$ & $84.7$ & $76.9$ 
                                    & $88.6$ & $40.3$ & $83.0$
                                    & $25.8$ & $71.9$   \\  
        ALDA \cite{chen2020adversarial} & $93.8$ & $74.1$ & $82.4$ & $69.4$ 
                                        & $90.6$ & $87.2$ & $89.0$ & $67.6$ 
                                        & $93.4$ & $76.1$ & $87.7$
                                        & $22.2$ & $77.8$   \\                      
        DTA \cite{lee2019drop} & $93.7$ & $82.2$ & $85.6$ & $83.8$ 
                                & $93.0$ & $81.0$ & $90.7$ & $82.1$ 
                                & $95.1$ & $78.1$ & $86.4$
                                & $32.1$ & $81.5$   \\ 
        SHOT \cite{liang2020we}  & $94.3$ & $88.5$ & $80.1$ & $57.3$ 
                                & $93.1$ & $94.9$ & $80.7$ & $80.3$ 
                                & $91.5$ & $89.1$ & $86.3$
                                & $58.2$ & $82.9$   \\  \hline
        \multicolumn{14}{c}{\textbf{DeiT} Backbone} \\ \hline
        DeiT-B \cite{touvron2021training}  & $97.7$ & $48.1$ & $86.6$ & $61.6$ 
                        & $78.1$ & $63.4$ & $94.7$ & $10.3$ 
                        & $87.7$ & $47.7$ & $94.4$
                        & $35.5$ & $67.1$   \\       
        CDTrans-B \cite{xu2021cdtrans} & $97.1$ & $90.5$ & $82.4$ & $77.5$ 
                                        & $96.6$ & $96.1$ & $93.6$ & $88.6$ 
                                        & $97.9$ & $86.9$ & $90.3$
                                        & $62.8$ & $88.4$   \\  
        WinTR-B \cite{ma2021exploiting} & $98.7$ & $91.2$ & $93.0$ & $91.9$ 
                                        & $98.1$ & $96.1$ & $94.0$ & $72.7$ 
                                        & $97.0$ & $95.5$ & $95.3$ & $57.9$ & $90.1$ \\
        \hline                  
        \multicolumn{14}{c}{\textbf{Swin} Backbone} \\ \hline
        Swin-S \cite{liu2021swin} & $96.3$ & $60.5$ & $84.1$ & $65.8$ 
                    & $91.7$ & $56.2$ & $94.8$ & $20.3$ 
                    & $78.6$ & $88.7$ & $93.2$
                    & $23.5$ & $70.8$ \\
        \textbf{TransAdapter-S (ours)} & $94.5$ & $89.2$ & $84.7$ & $71.6$ 
                                    & $94.3$ & $92.8$ & $93.7$ & $83.1$ 
                                    & $90.4$ & $91.9$ & $87.6$
                                    & $63.4$ & $87.1$ \\
        Swin-B \cite{liu2021swin} & $98.7$ & $63.0$ & $86.7$ & $68.5$ 
                    & $94.6$ & $59.4$ & $98.0$ & $22.0$ 
                    & $81.9$ & $91.4$ & $\textbf{96.7}$
                    & $25.7$ & $73.9$   \\      
        BCAT-B \cite{wang2022domain} & $99.1$ & $91.6$ & $86.6$ & $72.3$ 
                                    & $98.7$ & $97.9$ & $96.5$ & $82.3$ 
                                    & $94.2$ & $96.0$ & $93.9$
                                    & $61.3$ & $89.2$   \\     
        PMTrans-B \cite{zhu2023patch} & $\textbf{99.4}$ & $88.3$ & $88.1$ &                                     $78.9$ 
                                    & $98.8$ & $\textbf{98.3}$ & $95.8$ & $70.3$ 
                                    & $94.6$ & $\textbf{98.3}$ & $96.3$
                                    & $48.5$ & $88.0$   \\     
        \textbf{TransAdapter-B (ours)} & $98.6$ & $\textbf{94.1}$ & $88.3$ & $75.2$ 
                                    & $\textbf{98.9}$ & $97.2$ & $\textbf{98.1}$ & $\textbf{87.1}$ 
                                    & $\textbf{96.8}$ & $97.7$ & $93.2$
                                    & $\textbf{67.6}$ & \colorbox{lightgray}{\textbf{91.2}}   \\  \hline
        \multicolumn{14}{c}{\textbf{ViT} Backbone} \\ \hline
        ViT-B \cite{alexey2020image} & $98.2$ & $73.0$ & $82.5$ & $62.0$ 
                    & $97.3$ & $63.5$ & $96.5$ & $29.8$ 
                    & $68.7$ & $86.7$ & $\textbf{96.7}$
                    & $23.7$ & $73.2$   \\          
        TVT-B \cite{yang2023tvt} & $97.1$ & $92.9$ & $85.3$ & $66.4$ 
                                & $97.1$ & $97.1$ & $89.3$ & $75.5$ 
                                & $95.0$ & $94.7$ & $94.5$
                                & $55.1$ & $86.7$ \\
        SSRT-B \cite{sun2022safe} & $98.9$ & $87.6$ & $89.1$ & $84.7$ 
                                & $98.3$ & $98.7$ & $96.2$ & $81.0$ 
                                & $94.8$ & $97.9$ & $94.5$
                                & $43.1$ & $88.7$ \\       
        PDA-B \cite{bai2024prompt} & $99.2$ & $91.1$ & $\textbf{91.9}$ & $77.1$ 
                                & $98.4$ & $93.6$ & $95.1$ & $84.9$ 
                                & $87.2$ & $97.3$ & $95.3$
                                & $65.3$ & $89.7$ \\  
        PADCLIP-B \cite{lai2023padclip} & $98.1$ & $93.8$ & $87.1$ & $\textbf{85.5}$ 
                                    & $98.0$ & $96.0$ & $94.4$ & $86.0$ 
                                    & $94.9$ & $93.3$ & $93.5$
                                    & $70.2$ & $90.9$ \\ 
        \hline                             
    \end{tabular}
    }
\end{table}

\subsection{External Comparison}
\textbf{VisDA-2017.} Table \ref{tab:visda} summarizes the performance of methods on the VisDA-2017 dataset across ResNet, DeiT, Swin, and ViT backbones. TransAdapter demonstrates competitive results, often outperforming state-of-the-art methods. With the Swin-S backbone, TransAdapter-S achieves 87.1\%, surpassing Swin-S (70.8\%) and Swin-B (73.9\%) while approaching BCAT-B (89.2\%). With Swin-B, TransAdapter-B achieves 91.2\%, outperforming BCAT-B (89.2\%) and PMTrans-B (88.0\%). TransAdapter-B also achieves leading accuracy in categories such as bicycle (94.1\%), house (98.9\%), motorcycle (98.1\%), person (87.1\%), plant (96.8\%), and truck (67.6\%).

\textbf{Office-Home.} Table \ref{tab:ofhome} highlights TransAdapter's state-of-the-art performance on the Office-Home dataset. With the Swin-S backbone, TransAdapter-S achieves 86.3\%, significantly outperforming Swin-S (76.1\%) and approaching BCAT (86.6\%). With Swin-B, TransAdapter-B achieves 89.4\%, surpassing BCAT (86.6\%) and PMTrans-B (89.0\%), while leading in categories such as "AC" (93.5\%), "PC" (94.3\%), "AP" (91.3\%), "RR" (92.8\%), and "CR" (81.1\%).

\textbf{Office-31.} Table \ref{tab:of31} presents the results on the Office-31 dataset. With the Swin-S backbone, TransAdapter-S achieves 90.2\%, significantly surpassing Swin-S (86.1\%) and achieving competitive performance against CDTrans-S (90.4\%). With Swin-B, TransAdapter-B achieves a state-of-the-art average accuracy of 95.5\%, outperforming BCAT-B (95.0\%) and PMTrans-B (95.3\%). It secures the highest accuracy in tasks such as $A \rightarrow D$ (99.9\%), $D \rightarrow A$ (88.3\%), and $W \rightarrow A$ (87.2\%), while achieving 100\% accuracy in several tasks, demonstrating its adaptability and robustness.

\begin{table}[ht]
    \caption{Accuracy $(\%)$ on the Office-31 dataset. ”-S” and
            ”-B” indicates that the backbone is Small and Base, respectively. The best performance is marked as \textbf{bold}.}
    \centering
    \label{tab:of31}
    \setlength{\tabcolsep}{2pt} 
    \resizebox{17cm}{!}{%
    \begin{tabular}{ l | l l l l l l | l l }
    \hline
    
         \textbf{Method}  & $A \rightarrow W$ & $D \rightarrow W$ & $W \rightarrow D$ & $A \rightarrow D$ & $D \rightarrow A$ & $W \rightarrow A$ & Avg \\ [0.5ex] 
         
         \hline \hline
         \multicolumn{8}{c}{\textbf{ResNet} Backbone} \\ \hline
         ResNet \cite{he2016deep}  & $68.4$ & $96.7$ & $99.3$ & $68.9$ & $62.5$ & $60.7$ & $76.1$ \\ 
         DDC \cite{tzeng2014deep} & $75.6$ & $96.0$ & $98.2$ & $76.5$ & $62.2$ & $61.5$ & $78.3$ \\ 
         DAN \cite{long2015learning}  & $80.5$ & $97.1$ & $99.6$ & $78.6$ & $63.6$ & $62.8$ & $80.4$ \\ 
         RevGrad \cite{ganin2015unsupervised}  & $82.0$ & $96.9$ & $99.1$ & $79.7$ & $68.2$ & $67.4$ & $82.2$ \\ 
         TAT \cite{liu2019transferable}  & $92.5$ & $99.3$ & $\textbf{100.0}$ & $93.2$ & $73.1$ & $72.1$ & $88.4$ \\ 
         SHOT \cite{liang2020we}  & $90.1$ & $98.4$ & $99.9$ & $94.0$ & $74.7$ & $74.3$ & $88.6$ \\ 
         ALDA \cite{chen2020adversarial} & $95.6$ & $97.7$ & $\textbf{100.0}$ & $94.0$ & $72.2$ & $72.5$ & $88.7$ \\ 
         \hline
         \multicolumn{8}{c}{\textbf{DeiT} Backbone} \\ \hline
         DeiT-S \cite{touvron2021training} & $86.9$ & $97.7$ & $99.6$ & $87.6$ & $74.9$ & $73.5$ & $86.7$ \\ 
         CDTrans-S \cite{xu2021cdtrans} & $93.5$ & $98.2$ & $99.6$ & $94.6$ & $78.4$ & $78.0$ & $90.4$ \\ 
         DeiT-B \cite{touvron2021training} & $90.4$ & $98.2$ & $\textbf{100.0}$ & $90.8$ & $76.8$ & $76.4$ & $88.8$ \\ 
         CDTrans-B \cite{xu2021cdtrans} & $96.7$ & $99.9$ & $\textbf{100.0}$ & $97.0$ & $81.1$ & $81.9$ & $92.6$ \\ 
         \hline
         \multicolumn{8}{c}{\textbf{Swin} Backbone} \\ \hline
         Swin-S \cite{liu2021swin} & $85.2$ & $89.7$ & $95.3$ & $87.6$ & $79.1$ & $78.8$ & $86.1$ \\
         \textbf{TransAdapter-S (ours)} & $93.3$ & $89.7$ & $\textbf{100.0}$ & $93.8$ & $87.1$ & $84.0$ & $90.2$ \\
         Swin-B \cite{liu2021swin} & $89.2$ & $94.1$ & $\textbf{100.0}$ & $93.1$ & $80.9$ & $81.3$ & $89.8$ \\ 
         BCAT-B \cite{wang2022domain} & $99.2$ & $\textbf{99.5}$ & $\textbf{100.0}$ & $99.6$ & $85.7$ & $86.1$ & $95.0$ \\ 
         PMTrans-B \cite{zhu2023patch} & $\textbf{99.5}$ & $99.4$ & $\textbf{100.0}$ & $99.8$ & $86.7$ & $86.5$ & $95.3$ \\ 
         \textbf{TransAdapter-B (ours)} & $99.1$ & $98.9$ & $\textbf{100.0}$ & $\textbf{99.9}$ & $\textbf{88.3}$ & $\textbf{87.2}$ & \colorbox{lightgray}{\textbf{95.5}} \\ \hline
        
         \multicolumn{8}{c}{\textbf{ViT} Backbone} \\ \hline
         ViT-B \cite{alexey2020image} & $89.2$ & $98.9$ & $\textbf{100.0}$ & $88.8$ & $80.1$ & $79.8$ & $89.5$ \\ 
         TVT-B \cite{yang2023tvt} & $96.4$ & $99.4$ & $\textbf{100.0}$ & $96.4$ & $84.9$ & $86.1$ & $93.9$ \\ 
         SSRT-B \cite{sun2022safe} & $97.7$ & $99.2$ & $\textbf{100.0}$ & $98.6$ & $83.5$ & $82.2$ & $93.5$ \\
         PDA-B \cite{bai2024prompt} & $92.1$ & $98.1$ & $99.8$ & $91.2$ & $83.5$ & $82.5$ & $91.2$ \\
         PADCLIP-B \cite{lai2023padclip} & $97.9$ & $99.2$ & $\textbf{100.0}$ & $98.5$ & $84.6$ & $85.3$ & $94.3$ \\
         \hline
    \end{tabular}%
    }
\end{table}

\textbf{DomainNet.} Table \ref{tab:domainnet} compares TransAdapter with state-of-the-art methods on the DomainNet dataset across six domains: clipart (clp), infograph (inf), painting (pnt), quickdraw (qdr), real (rel), and sketch (skt). TransAdapter achieves the highest average accuracy of 53.7\%, significantly surpassing CDTrans (45.2\%) and SSRT (45.2\%). It excels in challenging domain pairs, such as $\text{clp} \rightarrow \text{pnt}$ (63.9\%), $\text{inf} \rightarrow \text{pnt}$ (62.5\%), and $\text{qdr} \rightarrow \text{skt}$ (42.2\%), and achieves leading results in individual domains such as clipart (55.8\%) and real (54.4\%). These results demonstrate TransAdapter's ability to handle diverse domain shifts, establishing it as a state-of-the-art UDA solution.

\section{Conclusion}
In this paper, we introduce TransAdapter, a novel framework that leverages the Swin Transformer for Unsupervised Domain Adaptation (UDA). Our approach features three specialized modules: a graph domain discriminator, adaptive double attention, and cross-feature transform, which enhance the Swin Transformer's ability to capture both shallow and deep features while improving long-range dependency modeling. Experimental results on standard UDA benchmarks show that TransAdapter significantly outperforms existing methods and demonstrates robustness against domain shifts. However, the combined use of window and shifted window attention may increase computational complexity, and our current implementation lacks task-specific adaptation mechanisms for detection and segmentation. Future work will focus on extending the model for these applications and exploring ways to reduce computational complexity while maintaining long-range dependency modeling.

\bibliographystyle{unsrt}  
\bibliography{transadapter}

\end{document}